\pdfoutput=1

\documentclass[11pt]{article}

\usepackage[]{EMNLP2022}

\usepackage{times}
\usepackage{latexsym}

\usepackage[T1]{fontenc}

\usepackage[utf8]{inputenc}

\usepackage{microtype}

\usepackage{inconsolata}

\usepackage{graphicx}
\usepackage{bm}
\usepackage{amsmath}
\usepackage{amssymb}
\usepackage{multirow}
\usepackage{booktabs}
\usepackage{amsthm}
\usepackage{subfig}
\DeclareMathOperator*{\argmax}{argmax}

\title{Viterbi Decoding of Directed Acyclic Transformer for Non-Autoregressive Machine Translation}

\author{Chenze Shao$^{1,2}$, Zhengrui Ma$^{1,2}$, Yang Feng$^{1,2}$\thanks{\ \ Corresponding author: Yang Feng}\\
$^{1}$ Key Laboratory of Intelligent Information Processing\\
Institute of Computing Technology, Chinese Academy of Sciences\\
$^{2}$ University of Chinese Academy of Sciences\\
{ \{\href{mailto:shaochenze18z@ict.ac.cn}{shaochenze18z}, \href{mailto:mazhengrui21b@ict.ac.cn}{mazhengrui21b}, \href{mailto:fengyang@ict.ac.cn}{fengyang}\}@ict.ac.cn}}

\begin{document}
\maketitle
\begin{abstract}
Non-autoregressive models achieve significant decoding speedup in neural machine translation but lack the ability to capture sequential dependency. Directed Acyclic Transformer (DA-Transformer) was recently proposed to model sequential dependency with a directed acyclic graph. Consequently, it has to apply a sequential decision process at inference time, which harms the global translation accuracy. In this paper, we present a Viterbi decoding framework for DA-Transformer, which guarantees to find the joint optimal solution for the translation and decoding path under any length constraint. Experimental results demonstrate that our approach consistently improves the performance of DA-Transformer while maintaining a similar decoding speedup.\footnote{We implement our method on https://github.com/thu-coai/DA-Transformer.}

\end{abstract}

\section{Introduction}
Non-autoregressive translation \cite{gu2017non} models achieve a significant decoding speedup but suffer from performance degradation, which is mainly attributed to the multi-modality problem. Multi-modality refers to the scenario where the same source sentence may have multiple translations with a strong cross-correlation between target words. However, non-autoregressive models generally hold the conditional independence assumption on target words, which prevents them from capturing the multimodal target distribution.

Recently, Directed Acyclic Transformer \cite{huang2022directed} was proposed to model sequential dependency with a directed acyclic graph consisting of different decoding paths that enable the model to capture multiple translation modalities. Although it has been proven effective, it cannot directly find the most probable translation with the argmax operation. Therefore, DA-Transformer has to apply a sequential decision process at inference time, which harms the global translation accuracy.

In this paper, we propose a Viterbi decoding \cite{1054010} framework for DA-Transformer to improve the decoding accuracy. Using the Markov property of decoding path, we can apply Viterbi decoding to find the most probable path, conditioned on which we can generate the translation with argmax decoding. Then, we further improve this decoding algorithm to perform a simultaneous search for decoding paths and translations, which guarantees to find the joint optimal solution under any length constraint. After Viterbi decoding, we obtain a set of translations with different lengths and rerank them to obtain the final translation. We apply a length penalty term in the reranking process, which prevents the generation of empty translation \cite{stahlberg-byrne-2019-nmt} and enables us to control the translation length flexibly.

Experimental results on several machine translation benchmark tasks (WMT14 En$\leftrightarrow$De, WMT17 Zh$\leftrightarrow$En) show that our approach consistently improves the performance of DA-Transformer while maintaining a similar decoding speedup.

\section{Preliminaries: DA-Transformer}
\subsection{Model Architecture}

DA-Transformer is formed by a Transformer encoder and a directed acyclic decoder. The encoder and layers of the decoder are the same as vanilla Transformer \cite{vaswani2017attention}. On top of the decoder, the hidden states are organized as a directed acyclic graph, whose edges represent transition probabilities between hidden states.

Given a source sentence $X=\{x_1,\cdots,x_N\}$ and a target sentence $Y=\{y_1,\cdots,y_M\}$, the decoder length $L$ is set to $\lambda \cdot N$, where $\lambda$ is a hyper-parameter. The translation probability from $X$ to $Y$ is formulated as:
\begin{equation}
P_{\theta}(Y|X) = \sum_{A \in \Gamma}P_{\theta}(A|X)P_{\theta}(Y|X,A),
\end{equation}
where $A=\{a_1,\cdots,a_M\}$ is a translation path for the target sentence $Y$ and $a_i$ represents the position of word $y_i$ in the decoder. $\Gamma$ contains all possible translation paths with $1=a_1 < \cdots < a_M = L$.

The probability of translation path $A$ is formulated based on the Markov hypothesis:
\begin{equation}
\label{eq:a}
P_{\theta}(A|X) = \prod_{i=1}^{M-1}P_{\theta}(a_{i+1}|a_{i},X)=\prod_{i=1}^{M-1}{E}_{a_{i},a_{i+1}},
\end{equation}
where ${E} \in \mathbb R^{L\times L}$ is the transition matrix obtained by self-attention, and ${E}_{a_{i},a_{i+1}}$ represents the transition probability from position $a_{i}$ to position $a_{i+1}$. ${E}$ is masked by a lower triangular matrix to ensure that the translation path is acyclic.

Conditioned on $X$ and the translation path $A$, the translation probability of $Y$ is formulated as:
\begin{equation}
\label{eq:y}
P_{\theta}(Y|A, X)=\prod_{i=1}^{M}P_{\theta}(y_i|a_i,X),
\end{equation}
where $P_{\theta}(y_i|a_i,X)$ represents the translation probability of word $y_i$ on the position $a_i$ of decoder.

\subsection{Training and Inference}

The training objective of DA-Transformer is to maximize the log-likelihood $\log P_{\theta}(Y|X)$, which requires marginalizing all paths $A$. Using the Markov property of translation path, DA-Transformer employs dynamic programming to calculate the translation probability. Besides, it applies glancing training \cite{qian-etal-2021-glancing} with a hyper-parameter $\tau$ to promote the learning. 

During inference, the objective is to find the most probable translation $\argmax_{Y}P_{\theta}(Y|X)$. However, there is no known tractable decoding algorithm for this problem. \citet{huang2022directed} proposed three approximate decoding strategies to find high-probability translations. The intuitive strategy is greedy decoding, which sequentially takes the most probable transition as the decoding path and generates a translation according to the conditional probabilities. Lookahead decoding improves greedy decoding by taking the most probable combination of transition and prediction as follows:

\begin{equation}
y_{i}^*, a_{i}^*=\argmax_{y_i,a_i}P_{\theta}(y_{i}|a_{i},X) P_{\theta}(a_{i}|a_{i-1}, X).
\end{equation}
Beam search decoding is a more accurate method that merges the paths of the same prefix, which approximates the real translation probability and better represents the model's preference. Beam search can be optionally combined with an n-gram language model to improve the performance further. However, the speed of beam search is much lower than greedy and lookahead decoding. 

\section{Methodology}
This section presents a Viterbi decoding framework for DA-Transformer to improve decoding accuracy. We first develop a basic algorithm to find the optimal decoding path and then improve it to find the joint optimal solution of the translations and decoding paths. Finally, we introduce the technique to rerank the Viterbi decoding outputs.

\subsection{Optimal Decoding Path}
\label{sec:31}
Recall that the greedy decoding strategy sequentially takes the most probable transition as the decoding path, which may not be optimal since the greedy strategy does not consider long-term profits. In response to this problem, we propose a Viterbi decoding framework for DA-Transformer that guarantees to find the optimal decoding path $\argmax_{A}P_\theta(A|X)$ under any length constraint.

Specifically, we consider decoding paths of length $i$ that end in position $a_i\!=\!t$, and use $\alpha(i,t)$ to represent the maximum probability of these paths. By definition, we set the initial state $\alpha(1,1)\!=\!1$ and $\alpha(1,t>1)\!=\!0$. The Markov property of decoding paths enables us to sequentially calculate $\alpha(i,\cdot)$ from its previous step $\alpha(i-1,\cdot)$:
\begin{equation}
\label{eq:v}
\begin{aligned}
\alpha(i,t) &= \max_{t'} \alpha(i-1,t') \cdot {E}_{t,t'},\\
\psi(i,t) &= \argmax_{t'} \alpha(i-1,t')\cdot {E}_{t,t'},
\end{aligned}
\end{equation}
where $E$ is the transition matrix defined in Equation \ref{eq:a} and $\psi(i,t)$ is the backtracking index pointing to the previous position. After $L$ iterations, we obtain the score for every possible length, and then we can find the optimal length with the argmax function:

\begin{equation}
\label{eq:m}
M=\argmax_{i} \alpha(i,L).
\end{equation}
After determining the length $M$, we can trace the best decoding path along the backtracking index starting from $a_M=L$:
\begin{equation}
a_i = \psi(i+1,a_{i+1}).
\end{equation}
Finally, conditioning on the optimal path $A$, we can generate the translation with argmax decoding:
\begin{equation}
y_i = \argmax_{y_i} P_{\theta}(y_i|a_i,X).
\end{equation}

\subsection{Joint Optimal Solution}
\label{sec:32}
The decoding algorithm described above can be summarized as the following process:
\begin{equation}
\begin{aligned}
A^*&=\argmax_{A}P_\theta(A|X),\\
Y^*&=\argmax_{Y}P_\theta(Y|X,A^*).
\end{aligned}
\end{equation}
Even though the algorithm now finds the optimal decoding path, the translation on this path may have low confidence, resulting in a low joint probability $P_\theta(A,Y|X)$. We further improve the decoding algorithm to search for both decoding paths and translations, which guarantees to find the joint optimal solution:
\begin{equation}
A^*,Y^*=\argmax_{A,Y}P_\theta(A,Y|X).
\end{equation}

Notice that when the path $A$ is given, we can easily find the most probable translation $Y$ with argmax decoding. Let $Y^A$ denotes the argmax decoding result under path $A$, where $y^{a_i}_i= \argmax_{y_i} P_{\theta}(y_i|a_i,X)$ is the $i$-th word of $Y^A$. Then we can simplify our objective with $Y^A$:
\begin{equation}
\label{eq:ay}
\begin{aligned}
&\max_{A,Y} P_{\theta}(A,\!Y|X)=\max_{A,Y} P_{\theta}(A|X)P_{\theta}(Y|X,\!A)\\
&=\max_{A}(P_{\theta}(A|X)\max_{Y}P_{\theta}(Y|X,\!A))\\
&=\max_{A} P_{\theta}(A|X)P_{\theta}(Y^A|X,A)\\
&=\max_{A}P_{\theta}(y^{a_1}_1|a_1,\!X)\!\prod_{i=1}^{M-1}\!{E}_{a_{i},a_{i+1}}P_{\theta}(y^{a_{i+1}}_{i+1}|a_{i+1},\!X)\\
&=\max_{A}P_{\theta}(y^A_1|a_1, X)\prod_{i=1}^{M-1}{E'}_{a_{i},a_{i+1}},
\end{aligned}
\end{equation}
where we introduce a new transition matrix $E'$ with $E'_{a_{i},a_{i+1}}\!=\!{E}_{a_{i},a_{i+1}}P_{\theta}(y^{a_{i+1}}_{i+1}|a_{i+1}, X)$. Compared to $\max_{A} P_{\theta}(A|X)$, the major difference is the transition matrix $E'$, which considers both the transition probability and the prediction probability. Therefore, we can still apply the Viterbi decoding framework to find the optimal joint solution.

We use `Viterbi' to represent the Viterbi decoding algorithm proposed in section \ref{sec:31}, and use `Joint-Viterbi' to represent the improved algorithm in this section that finds the joint optimal solution. It is worth noting that Viterbi and Joint-Viterbi can be regarded as improvements to greedy decoding and lookahead decoding, respectively. Both greedy decoding and lookahead decoding consider the one-step probability and find the next token with $\argmax_{a_i} P_{\theta}(a_i|X,a_{i-1})$ and $\argmax_{y_i,a_i}P_{\theta}(y_{i}|a_{i},X) P_{\theta}(a_{i}|a_{i-1}, X)$, respectively. In comparison, Viterbi and Joint-Viterbi consider the whole decoding path and guarantee to find the global optimal solution $\argmax_{A}P_\theta(A|X)$ and $\argmax_{A,Y}P_\theta(A,Y|X)$, respectively.

\subsection{Reranking with Length Penalty}
After Viterbi decoding, we have a set of translations of different lengths that can be ranked to obtain the most probable one. However, argmax decoding is biased toward short translations and may even degenerate to an empty translation, as observed in \citet{stahlberg-byrne-2019-nmt}. 

To solve this problem, we introduce the hyper-parameter $\beta$ for length normalization in \citet{wu2016google} and modify Equation \ref{eq:m} to divide by the length penalty term:

\begin{equation}
M=\argmax_{i} \frac{\log\alpha(i,L)}{i^\beta}.
\end{equation}

By changing the length penalty $\beta$ to different values, we now have the flexibility to control the translation length with little additional overhead, which is another appealing feature of our approach. 

\section{Experiments}
\subsection{Settings}
\begin{table*}[th!]
\centering
\small
\begin{tabular}{lcccccccc}
\toprule
 \multirow{2}{*}{\textbf{Models}} & \multirow{2}{*}{\textbf{Iter}} &
 \multicolumn{2}{c}{\textbf{WMT14}} & \multicolumn{2}{c}{\textbf{WMT17}} &\textbf{Average}&\multirow{2}{*}{\textbf{Speedup}} &\\
  &&\textbf{En-De} & \textbf{De-En} & \textbf{En-Zh} & \textbf{Zh-En} &\textbf{Gap}\\
\midrule
 Transformer & $M$&27.67&31.84&35.05&24.26 &0&1.0$\times$\\
\cmidrule[0.6pt](lr){1-8}
DA-Transformer + Greedy  &$1$&26.06  & 30.69 & 33.29 &22.32  &1.62& 14.2$\times$\\
DA-Transformer + Viterbi  &$1$& 26.43$^\dagger$ & 30.84 & 33.25 &  22.58$^\dagger$&1.43& 13.3$\times$ \\
\cmidrule[0.6pt](lr){1-8}
DA-Transformer + Lookahead  &$1$&26.55  & 30.81 & 33.54 &22.68  &1.31& 14.0$\times$\\
DA-Transformer + Joint-Viterbi  &$1$& 26.89$^\dagger$ & 31.10$^\dagger$ & 33.65 &  23.24$^\dagger$&0.98& 13.2$\times$ \\
\bottomrule
\end{tabular}
\caption{Results on WMT14 En$\leftrightarrow$De and WMT17 Zh$\leftrightarrow$En. $M$ is the length of the target sentence.  `Iter' means the number of decoding iterations. The speedup is evaluated on WMT14 En-De test set with a batch size of 1. $\dagger$ means significantly better than the baseline model ($p < 0.05$). We use the statistical significance test with paired bootstrap resampling \cite{koehn-2004-statistical}.}
\label{tab:main-rst}
\end{table*}

We conduct experiments on WMT14 English$\leftrightarrow$German (En$\leftrightarrow$De, 4.5M pairs) and WMT17 Chinese$\leftrightarrow$English (Zh$\leftrightarrow$En, 20M pairs). These datasets are all encoded into subword units \cite{sennrich-etal-2016-neural}. We use the same preprocessed data and train/dev/test splits as \citet{Kasai2020NonautoregressiveMT}. The translation quality is evaluated with sacreBLEU \cite{post-2018-call} for WMT17 En-Zh and tokenized BLEU \cite{papineni2002bleu} for other benchmarks. We use GeForce RTX $3090$ to train models and measure translation latency. Our models are implemented based on the open-source toolkit of fairseq \cite{ott2019fairseq}.

We strictly follow the hyper-parameter settings of \citet{huang2022directed} to reimplement DA-Transformer. We adopt Transformer-base \cite{vaswani2017attention} as the model architecture. We set dropout to 0.1, weight decay to 0.01, and label smoothing to 0.1 for regularization. We use $\lambda=8$ for the graph size and linearly anneal $\tau$ from 0.5 to 0.1 for the glancing training. For fair comparisons, we tune the length penalty in $[0.95,1.05]$ to obtain a similar translation length as lookahead. We train all models for 300K steps, where each batch contains approximately 64K source tokens. All models are optimized by Adam \cite{DBLP:journals/corr/KingmaB14} with $\beta=(0.9,0.999)$ and $\epsilon=10^{-8}$. The learning rate warms up to $5\cdot 10^{-4}$ and then begins to anneal it after 10K steps with the inverse square-root schedule. We calculate the validation BLEU scores every epoch and obtain the final model by taking an average of the best five checkpoints. 

\subsection{Main Results}
As shown in Table \ref{tab:main-rst}, both Viterbi and Joint-Viterbi improve over their corresponding baseline. Joint-Viterbi achieves the best performance, which outperforms the previous lookahead strategy by 0.33 BLEU. Besides, it is worth noting that the Viterbi decoding process is highly parallelizable, which does not bring much overhead in the decoding and only reduces the speedup by less than $1\times$. 

\subsection{Results with Knowledge Distillation}
In this section, we evaluate the performance of our method with sequence-level knowledge distillation \cite{hinton2015distilling,kim-rush-2016-sequence}, where the target side of the training set is replaced by the output of an autoregressive teacher model. Experimental results in Table \ref{tab:kd} show that the differences between decoding strategies are relatively small.

Intuitively, we attribute this phenomenon to the improvement of model confidence. As knowledge distillation reduces the multi-modality of the dataset \cite{Zhou2020Understanding,pmlr-v119-sun20c}, the model may become more confident in predicting target sentences, which makes the greedy strategy more likely to reach the optima. To verify this, we measure the average entropy of transition and prediction probabilities and evaluate the percentage of lookahead outputs that match the optima $\argmax_{A,Y}P(A,Y|X)$ under their length. As Table \ref{tab:wokd} shows, DA-Transformer with distillation has smaller entropies and a larger percentage of optimal translations, which confirms our intuition.

\begin{table}[t]
\centering
\resizebox{\columnwidth}{!}{
\begin{tabular}{ccccc}
\toprule
Method&Greedy& Lookahead & Viterbi& Joint-Viterbi\\
\midrule
BLEU&26.81&26.91&26.88&27.03\\
\bottomrule
\end{tabular}
}
\caption{Results with knowledge distillation on WMT14 En-De test set.}
\label{tab:kd}
\end{table}

\begin{table}[t]
\small
\centering
\begin{tabular}{cccc}
\toprule
Metric&T-Entropy&P-Entropy& Percentage  \\
\midrule
w/o kd&1.088&1.892&59.6\%\\
w/ kd&0.998&0.601&70.1\%\\
\bottomrule
\end{tabular}
\caption{Statistics of DA-Transformer on WMT14 En-De test set. `kd' means knowledge distillation. `T-' means transition and `P-' means prediction.}
\label{tab:wokd}
\end{table}
\subsection{Probability Analysis}
\label{sec:44}
Recall that the decoding objective is to find the most probable translation $\argmax_{Y}P(Y|X)$, while our approach finds the joint solution $\argmax_{A,Y}P(A,Y|X)$. Although there is a gap between them, we argue that optimizing the joint probability helps us achieve higher translation probability. To prove it, we collect the outputs of lookahead decoding and Joint-Viterbi on WMT14 En-De test set and compute their probabilities $P(Y|X)$ by dynamic programming. We then calculate the average log probability of each decoding strategy, and also evaluate the percentage of translations that one strategy obtains a larger probability than another. As Table \ref{tab:prob} shows, Joint-Viterbi outperforms lookahead decoding by a large margin, indicating that we can obtain a higher average translation probability by optimizing the joint probability.

\begin{table}[t]
\small
\centering
\begin{tabular}{ccc}
\toprule
Method&Lookahead&Joint-Viterbi\\
\midrule
Log-prob&$-$4.39&$-$4.14\\
Percentage&24.4\%&41.6\%\\
\bottomrule
\end{tabular}
\caption{Probability analysis of on Lookahead and Joint-Viterbi decoding on WMT14 En-De test set.}
\label{tab:prob}
\end{table}

\subsection{Effect of Length Penalty}
\begin{figure}[h]
  \begin{center}
    \includegraphics[width=1\columnwidth]{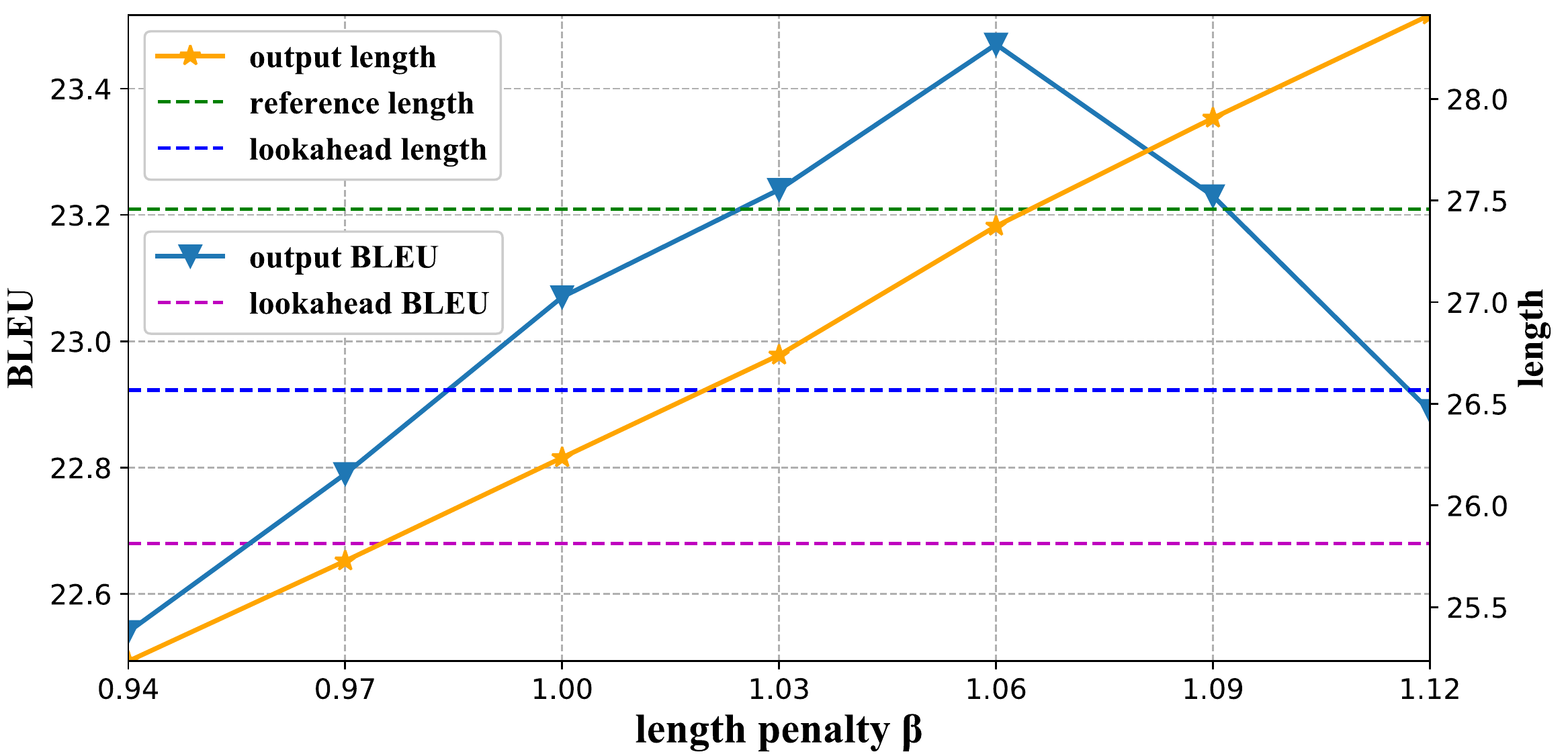}
    \caption{The effect of length penalty $\beta$ measured on WMT17 Zh-En test set.}
    \label{fig:length}
  \end{center}
\end{figure}

Viterbi decoding is capable of flexibly controlling the output length with the length penalty $\beta$. To show the effect of the length penalty, we change the value of $\beta$ in Joint-Viterbi to decode the WMT17 Zh-En test set and report the corresponding BLEU scores and average output lengths in Figure \ref{fig:length}. It shows that the length penalty can almost linearly control the output length, which can help us obtain satisfactory translations. Generally, Viterbi decoding can obtain better performance when the output length is closer to the reference length. If there is no length penalty, only finding outputs with the maximum joint probability will break the translation quality with extremely small output lengths.

\section{Related Works}
Most non-autoregressive models can directly find the most probable output with argmax decoding, which is the fastest decoding algorithm. However, models of this type usually suffer from the multi-modality problem \cite{gu2017non}, leading to severe performance degradation. A relatively more accurate method is noisy parallel decoding, which requires generating multiple translation candidates and greatly increases the amount of computation.

Many efforts have been made to address the multi-modality problem, including latent models \cite{kaiser2018fast,Ma_2019,Shu2020LatentVariableNN,bao-etal-2021-non,bao-etal-2022-textit}, alignment-based models \cite{gu2017non,ran2019guiding,song-etal-2021-alignart}, and better training objectives \cite{shao-etal-2019-retrieving,DBLP:conf/aaai/ShaoZFMZ20,shan2021modeling,Aligned,Du2021OAXE,DBLP:journals/corr/abs-2106-08122}. However, these techniques are still not powerful enough, which heavily rely on knowledge distillation \cite{kim-rush-2016-sequence}.

Some researchers seek iterative decoding approaches to improve translation quality. Work in this area includes semi-autoregressive decoding \cite{wang2018semi}, iterative refinement \cite{lee2018deterministic}, mask-predict decoding \cite{ghazvininejad2019maskpredict}, Levenshtein Transformer \cite{gu2019levenshtein}, multi-thread decoding \cite{ran-etal-2020-learning}, Imputer \cite{saharia-etal-2020-non}, and rewriting \cite{geng-etal-2021-learning}. Although their translations are of better quality, they are criticized for being slow at inference time \cite{DBLP:conf/iclr/Kasai0PCS21}.

Recently, latent alignment models like CTC \cite{libovicky2018end,saharia-etal-2020-non} and DA-Transformer \cite{huang2022directed} achieved impressive performance and received a lot of attention. Beam search is an useful decoding strategy for latent alignment models \cite{kasner2020improving,gu2020fully,zheng2021duplex,shao2022one,huang2022directed,nmla}. It brings considerable improvements but also reduces the decoding speed. 

Viterbi decoding has also been used in non-autoregressive models. In CRF-based NAT models, Viterbi decoding is applied to find the most probable output \cite{NIPS2019_8566,pmlr-v119-sun20c}.

\section{Conclusion}
The current decoding strategies of DA-Transformer need to apply a sequential decision process, which harms the global translation accuracy. In this paper, we propose a Viterbi decoding framework for DA-Transformer to find the joint optimal solution of the translation and decoding path and further demonstrate its effectiveness on multiple benchmarks.
\section{Acknowledgement}
We thank the anonymous reviewers for their insightful comments. We thank Fei Huang for helping us open source code.
\section*{Limitations}
The major limitation of our method is that it cannot find the most probable translation $\argmax_{Y}P(Y|X)$ but alternatively finds the joint optimal solution $\argmax_{A,Y}P(A,Y|X)$. However, as we show in section \ref{sec:44}, outputs with higher joint probability usually also have higher translation probability, suggesting that optimizing the joint probability is helpful.

Another limitation is that the improvements of our method are smaller in the knowledge distillation setting. However, the main advantage of DA-Transformer is that it does not heavily rely on knowledge distillation and achieves superior performance on raw data, which makes the impact of this limitation small.

\bibliography{custom}

\begin{thebibliography}{46}
\expandafter\ifx\csname natexlab\endcsname\relax\def\natexlab#1{#1}\fi

\bibitem[{Bao et~al.(2021)Bao, Huang, Xiao, Wang, Dai, and
  Chen}]{bao-etal-2021-non}
Yu~Bao, Shujian Huang, Tong Xiao, Dongqi Wang, Xinyu Dai, and Jiajun Chen.
  2021.
\newblock \href {https://doi.org/10.18653/v1/2021.naacl-main.458}
  {Non-autoregressive translation by learning target categorical codes}.
\newblock In \emph{Proceedings of the 2021 Conference of the North American
  Chapter of the Association for Computational Linguistics: Human Language
  Technologies}, pages 5749--5759, Online. Association for Computational
  Linguistics.

\bibitem[{Bao et~al.(2022)Bao, Zhou, Huang, Wang, Qian, Dai, Chen, and
  Li}]{bao-etal-2022-textit}
Yu~Bao, Hao Zhou, Shujian Huang, Dongqi Wang, Lihua Qian, Xinyu Dai, Jiajun
  Chen, and Lei Li. 2022.
\newblock \href {https://doi.org/10.18653/v1/2022.acl-long.575} {{latent-GLAT}:
  Glancing at latent variables for parallel text generation}.
\newblock In \emph{Proceedings of the 60th Annual Meeting of the Association
  for Computational Linguistics (Volume 1: Long Papers)}, pages 8398--8409,
  Dublin, Ireland. Association for Computational Linguistics.

\bibitem[{Du et~al.(2021)Du, Tu, and Jiang}]{Du2021OAXE}
Cunxiao Du, Zhaopeng Tu, and Jing Jiang. 2021.
\newblock Order-agnostic cross entropy for non-autoregressive machine
  translation.
\newblock In \emph{ICML}.

\bibitem[{Geng et~al.(2021)Geng, Feng, and Qin}]{geng-etal-2021-learning}
Xinwei Geng, Xiaocheng Feng, and Bing Qin. 2021.
\newblock \href {https://doi.org/10.18653/v1/2021.emnlp-main.265} {Learning to
  rewrite for non-autoregressive neural machine translation}.
\newblock In \emph{Proceedings of the 2021 Conference on Empirical Methods in
  Natural Language Processing}, pages 3297--3308, Online and Punta Cana,
  Dominican Republic. Association for Computational Linguistics.

\bibitem[{Ghazvininejad et~al.(2020)Ghazvininejad, Karpukhin, Zettlemoyer, and
  Levy}]{Aligned}
Marjan Ghazvininejad, Vladimir Karpukhin, Luke Zettlemoyer, and Omer Levy.
  2020.
\newblock Aligned cross entropy for non-autoregressive machine translation.
\newblock In \emph{ICML}.

\bibitem[{Ghazvininejad et~al.(2019)Ghazvininejad, Levy, Liu, and
  Zettlemoyer}]{ghazvininejad2019maskpredict}
Marjan Ghazvininejad, Omer Levy, Yinhan Liu, and Luke Zettlemoyer. 2019.
\newblock \href {https://www.aclweb.org/anthology/D19-1633} {Mask-predict:
  Parallel decoding of conditional masked language models}.
\newblock In \emph{Proceedings of the 2019 Conference on Empirical Methods in
  Natural Language Processing and the 9th International Joint Conference on
  Natural Language Processing (EMNLP-IJCNLP)}, pages 6112--6121.

\bibitem[{Gu et~al.(2018)Gu, Bradbury, Xiong, Li, and Socher}]{gu2017non}
Jiatao Gu, James Bradbury, Caiming Xiong, Victor~O.K. Li, and Richard Socher.
  2018.
\newblock \href {https://openreview.net/forum?id=B1l8BtlCb} {Non-autoregressive
  neural machine translation}.
\newblock In \emph{International Conference on Learning Representations}.

\bibitem[{Gu and Kong(2020)}]{gu2020fully}
Jiatao Gu and Xiang Kong. 2020.
\newblock \href {http://arxiv.org/abs/2012.15833} {Fully non-autoregressive
  neural machine translation: Tricks of the trade}.

\bibitem[{Gu et~al.(2019)Gu, Wang, and Zhao}]{gu2019levenshtein}
Jiatao Gu, Changhan Wang, and Junbo Zhao. 2019.
\newblock \href
  {https://proceedings.neurips.cc/paper/2019/file/675f9820626f5bc0afb47b57890b466e-Paper.pdf}
  {Levenshtein transformer}.
\newblock In \emph{Advances in Neural Information Processing Systems},
  volume~32. Curran Associates, Inc.

\bibitem[{Hinton et~al.(2015)Hinton, Vinyals, and Dean}]{hinton2015distilling}
Geoffrey Hinton, Oriol Vinyals, and Jeff Dean. 2015.
\newblock Distilling the knowledge in a neural network.
\newblock \emph{arXiv preprint arXiv:1503.02531}.

\bibitem[{Huang et~al.(2022)Huang, Zhou, Liu, Li, and
  Huang}]{huang2022directed}
Fei Huang, Hao Zhou, Yang Liu, Hang Li, and Minlie Huang. 2022.
\newblock Directed acyclic transformer for non-autoregressive machine
  translation.
\newblock In \emph{Proceedings of the 39th International Conference on Machine
  Learning, {ICML} 2022}.

\bibitem[{Kaiser et~al.(2018)Kaiser, Bengio, Roy, Vaswani, Parmar, Uszkoreit,
  and Shazeer}]{kaiser2018fast}
Lukasz Kaiser, Samy Bengio, Aurko Roy, Ashish Vaswani, Niki Parmar, Jakob
  Uszkoreit, and Noam Shazeer. 2018.
\newblock \href {http://proceedings.mlr.press/v80/kaiser18a.html} {Fast
  decoding in sequence models using discrete latent variables}.
\newblock In \emph{Proceedings of the 35th International Conference on Machine
  Learning}, volume~80 of \emph{Proceedings of Machine Learning Research},
  pages 2390--2399. PMLR.

\bibitem[{Kasai et~al.(2020)Kasai, Cross, Ghazvininejad, and
  Gu}]{Kasai2020NonautoregressiveMT}
Jungo Kasai, James Cross, Marjan Ghazvininejad, and Jiatao Gu. 2020.
\newblock Non-autoregressive machine translation with disentangled context
  transformer.
\newblock In \emph{ICML}.

\bibitem[{Kasai et~al.(2021)Kasai, Pappas, Peng, Cross, and
  Smith}]{DBLP:conf/iclr/Kasai0PCS21}
Jungo Kasai, Nikolaos Pappas, Hao Peng, James Cross, and Noah~A. Smith. 2021.
\newblock \href {https://openreview.net/forum?id=KpfasTaLUpq} {Deep encoder,
  shallow decoder: Reevaluating non-autoregressive machine translation}.
\newblock In \emph{9th International Conference on Learning Representations,
  {ICLR} 2021, Virtual Event, Austria, May 3-7, 2021}. OpenReview.net.

\bibitem[{Kasner et~al.(2020)Kasner, Libovický, and
  Helcl}]{kasner2020improving}
Zdeněk Kasner, Jindřich Libovický, and Jindřich Helcl. 2020.
\newblock \href {http://arxiv.org/abs/2004.03227} {Improving fluency of
  non-autoregressive machine translation}.

\bibitem[{Kim and Rush(2016)}]{kim-rush-2016-sequence}
Yoon Kim and Alexander~M. Rush. 2016.
\newblock \href {https://doi.org/10.18653/v1/D16-1139} {Sequence-level
  knowledge distillation}.
\newblock In \emph{Proceedings of the 2016 Conference on Empirical Methods in
  Natural Language Processing}, pages 1317--1327, Austin, Texas. Association
  for Computational Linguistics.

\bibitem[{Kingma and Ba(2014)}]{DBLP:journals/corr/KingmaB14}
Diederik~P. Kingma and Jimmy Ba. 2014.
\newblock \href {http://arxiv.org/abs/1412.6980} {Adam: {A} method for
  stochastic optimization}.
\newblock \emph{CoRR}, abs/1412.6980.

\bibitem[{Koehn(2004)}]{koehn-2004-statistical}
Philipp Koehn. 2004.
\newblock \href {https://aclanthology.org/W04-3250} {Statistical significance
  tests for machine translation evaluation}.
\newblock In \emph{Proceedings of the 2004 Conference on Empirical Methods in
  Natural Language Processing}, pages 388--395, Barcelona, Spain. Association
  for Computational Linguistics.

\bibitem[{Lee et~al.(2018)Lee, Mansimov, and Cho}]{lee2018deterministic}
Jason Lee, Elman Mansimov, and Kyunghyun Cho. 2018.
\newblock \href {https://doi.org/10.18653/v1/D18-1149} {Deterministic
  non-autoregressive neural sequence modeling by iterative refinement}.
\newblock In \emph{Proceedings of the 2018 Conference on Empirical Methods in
  Natural Language Processing}, pages 1173--1182, Brussels, Belgium.
  Association for Computational Linguistics.

\bibitem[{Libovick{\'y} and Helcl(2018)}]{libovicky2018end}
Jind{\v{r}}ich Libovick{\'y} and Jind{\v{r}}ich Helcl. 2018.
\newblock \href {https://doi.org/10.18653/v1/D18-1336} {End-to-end
  non-autoregressive neural machine translation with connectionist temporal
  classification}.
\newblock In \emph{Proceedings of the 2018 Conference on Empirical Methods in
  Natural Language Processing}, pages 3016--3021, Brussels, Belgium.
  Association for Computational Linguistics.

\bibitem[{Ma et~al.(2019)Ma, Zhou, Li, Neubig, and Hovy}]{Ma_2019}
Xuezhe Ma, Chunting Zhou, Xian Li, Graham Neubig, and Eduard Hovy. 2019.
\newblock \href {https://doi.org/10.18653/v1/D19-1437} {{F}low{S}eq:
  Non-autoregressive conditional sequence generation with generative flow}.
\newblock In \emph{Proceedings of the 2019 Conference on Empirical Methods in
  Natural Language Processing and the 9th International Joint Conference on
  Natural Language Processing (EMNLP-IJCNLP)}, pages 4282--4292, Hong Kong,
  China. Association for Computational Linguistics.

\bibitem[{Ott et~al.(2019)Ott, Edunov, Baevski, Fan, Gross, Ng, Grangier, and
  Auli}]{ott2019fairseq}
Myle Ott, Sergey Edunov, Alexei Baevski, Angela Fan, Sam Gross, Nathan Ng,
  David Grangier, and Michael Auli. 2019.
\newblock fairseq: A fast, extensible toolkit for sequence modeling.
\newblock In \emph{Proceedings of NAACL-HLT 2019: Demonstrations}.

\bibitem[{Papineni et~al.(2002)Papineni, Roukos, Ward, and
  Zhu}]{papineni2002bleu}
Kishore Papineni, Salim Roukos, Todd Ward, and Wei-Jing Zhu. 2002.
\newblock \href {https://doi.org/10.3115/1073083.1073135} {{B}leu: a method for
  automatic evaluation of machine translation}.
\newblock In \emph{Proceedings of the 40th Annual Meeting of the Association
  for Computational Linguistics}, pages 311--318, Philadelphia, Pennsylvania,
  USA. Association for Computational Linguistics.

\bibitem[{Post(2018)}]{post-2018-call}
Matt Post. 2018.
\newblock \href {https://doi.org/10.18653/v1/W18-6319} {A call for clarity in
  reporting {BLEU} scores}.
\newblock In \emph{Proceedings of the Third Conference on Machine Translation:
  Research Papers}, pages 186--191, Brussels, Belgium. Association for
  Computational Linguistics.

\bibitem[{Qian et~al.(2021)Qian, Zhou, Bao, Wang, Qiu, Zhang, Yu, and
  Li}]{qian-etal-2021-glancing}
Lihua Qian, Hao Zhou, Yu~Bao, Mingxuan Wang, Lin Qiu, Weinan Zhang, Yong Yu,
  and Lei Li. 2021.
\newblock \href {https://doi.org/10.18653/v1/2021.acl-long.155} {Glancing
  transformer for non-autoregressive neural machine translation}.
\newblock In \emph{Proceedings of the 59th Annual Meeting of the Association
  for Computational Linguistics and the 11th International Joint Conference on
  Natural Language Processing (Volume 1: Long Papers)}, pages 1993--2003,
  Online. Association for Computational Linguistics.

\bibitem[{Ran et~al.(2020)Ran, Lin, Li, and Zhou}]{ran-etal-2020-learning}
Qiu Ran, Yankai Lin, Peng Li, and Jie Zhou. 2020.
\newblock \href {https://doi.org/10.18653/v1/2020.acl-main.277} {Learning to
  recover from multi-modality errors for non-autoregressive neural machine
  translation}.
\newblock In \emph{Proceedings of the 58th Annual Meeting of the Association
  for Computational Linguistics}, pages 3059--3069, Online. Association for
  Computational Linguistics.

\bibitem[{Ran et~al.(2021)Ran, Lin, Li, and Zhou}]{ran2019guiding}
Qiu Ran, Yankai Lin, Peng Li, and Jie Zhou. 2021.
\newblock \href {https://ojs.aaai.org/index.php/AAAI/article/view/17618}
  {Guiding non-autoregressive neural machine translation decoding with
  reordering information}.
\newblock In \emph{Thirty-Fifth {AAAI} Conference on Artificial Intelligence,
  {AAAI} 2021, Virtual Event, February 2-9, 2021}, pages 13727--13735. {AAAI}
  Press.

\bibitem[{Saharia et~al.(2020)Saharia, Chan, Saxena, and
  Norouzi}]{saharia-etal-2020-non}
Chitwan Saharia, William Chan, Saurabh Saxena, and Mohammad Norouzi. 2020.
\newblock \href {https://doi.org/10.18653/v1/2020.emnlp-main.83}
  {Non-autoregressive machine translation with latent alignments}.
\newblock In \emph{Proceedings of the 2020 Conference on Empirical Methods in
  Natural Language Processing (EMNLP)}, pages 1098--1108, Online. Association
  for Computational Linguistics.

\bibitem[{Sennrich et~al.(2016)Sennrich, Haddow, and
  Birch}]{sennrich-etal-2016-neural}
Rico Sennrich, Barry Haddow, and Alexandra Birch. 2016.
\newblock \href {https://doi.org/10.18653/v1/P16-1162} {Neural machine
  translation of rare words with subword units}.
\newblock In \emph{Proceedings of the 54th Annual Meeting of the Association
  for Computational Linguistics (Volume 1: Long Papers)}, pages 1715--1725,
  Berlin, Germany. Association for Computational Linguistics.

\bibitem[{Shan et~al.(2021)Shan, Feng, and Shao}]{shan2021modeling}
Yong Shan, Yang Feng, and Chenze Shao. 2021.
\newblock Modeling coverage for non-autoregressive neural machine translation.
\newblock In \emph{2021 International Joint Conference on Neural Networks
  (IJCNN)}, pages 1--8. IEEE.

\bibitem[{Shao and Feng(2022)}]{nmla}
Chenze Shao and Yang Feng. 2022.
\newblock Non-monotonic latent alignments for ctc-based non-autoregressive
  machine translation.
\newblock In \emph{Proceedings of NeurIPS 2022}.

\bibitem[{Shao et~al.(2019)Shao, Feng, Zhang, Meng, Chen, and
  Zhou}]{shao-etal-2019-retrieving}
Chenze Shao, Yang Feng, Jinchao Zhang, Fandong Meng, Xilin Chen, and Jie Zhou.
  2019.
\newblock \href {https://doi.org/10.18653/v1/P19-1288} {Retrieving sequential
  information for non-autoregressive neural machine translation}.
\newblock In \emph{Proceedings of the 57th Annual Meeting of the Association
  for Computational Linguistics}, pages 3013--3024, Florence, Italy.
  Association for Computational Linguistics.

\bibitem[{Shao et~al.(2021)Shao, Feng, Zhang, Meng, and
  Zhou}]{DBLP:journals/corr/abs-2106-08122}
Chenze Shao, Yang Feng, Jinchao Zhang, Fandong Meng, and Jie Zhou. 2021.
\newblock \href {https://doi.org/10.1162/coli_a_00421} {{Sequence-Level
  Training for Non-Autoregressive Neural Machine Translation}}.
\newblock \emph{Computational Linguistics}, 47(4):891--925.

\bibitem[{Shao et~al.(2022)Shao, Wu, and Feng}]{shao2022one}
Chenze Shao, Xuanfu Wu, and Yang Feng. 2022.
\newblock \href {https://doi.org/10.18653/v1/2022.naacl-main.277} {One
  reference is not enough: Diverse distillation with reference selection for
  non-autoregressive translation}.
\newblock In \emph{Proceedings of the 2022 Conference of the North American
  Chapter of the Association for Computational Linguistics: Human Language
  Technologies}, pages 3779--3791, Seattle, United States. Association for
  Computational Linguistics.

\bibitem[{Shao et~al.(2020)Shao, Zhang, Feng, Meng, and
  Zhou}]{DBLP:conf/aaai/ShaoZFMZ20}
Chenze Shao, Jinchao Zhang, Yang Feng, Fandong Meng, and Jie Zhou. 2020.
\newblock \href {https://aaai.org/ojs/index.php/AAAI/article/view/5351}
  {Minimizing the bag-of-ngrams difference for non-autoregressive neural
  machine translation}.
\newblock In \emph{The Thirty-Fourth {AAAI} Conference on Artificial
  Intelligence, {AAAI} 2020, New York, NY, USA, February 7-12, 2020}, pages
  198--205. {AAAI} Press.

\bibitem[{Shu et~al.(2020)Shu, Lee, Nakayama, and
  Cho}]{Shu2020LatentVariableNN}
Raphael Shu, Jason Lee, Hideki Nakayama, and Kyunghyun Cho. 2020.
\newblock Latent-variable non-autoregressive neural machine translation with
  deterministic inference using a delta posterior.
\newblock In \emph{AAAI}.

\bibitem[{Song et~al.(2021)Song, Kim, and Yoon}]{song-etal-2021-alignart}
Jongyoon Song, Sungwon Kim, and Sungroh Yoon. 2021.
\newblock \href {https://doi.org/10.18653/v1/2021.emnlp-main.1} {{A}lig{NART}:
  Non-autoregressive neural machine translation by jointly learning to estimate
  alignment and translate}.
\newblock In \emph{Proceedings of the 2021 Conference on Empirical Methods in
  Natural Language Processing}, pages 1--14, Online and Punta Cana, Dominican
  Republic. Association for Computational Linguistics.

\bibitem[{Stahlberg and Byrne(2019)}]{stahlberg-byrne-2019-nmt}
Felix Stahlberg and Bill Byrne. 2019.
\newblock \href {https://doi.org/10.18653/v1/D19-1331} {On {NMT} search errors
  and model errors: Cat got your tongue?}
\newblock In \emph{Proceedings of the 2019 Conference on Empirical Methods in
  Natural Language Processing and the 9th International Joint Conference on
  Natural Language Processing (EMNLP-IJCNLP)}, pages 3356--3362, Hong Kong,
  China. Association for Computational Linguistics.

\bibitem[{Sun et~al.(2019)Sun, Li, Wang, He, Lin, and Deng}]{NIPS2019_8566}
Zhiqing Sun, Zhuohan Li, Haoqing Wang, Di~He, Zi~Lin, and Zhihong Deng. 2019.
\newblock \href
  {http://papers.nips.cc/paper/8566-fast-structured-decoding-for-sequence-models.pdf}
  {Fast structured decoding for sequence models}.
\newblock In \emph{Advances in Neural Information Processing Systems 32}, pages
  3016--3026.

\bibitem[{Sun and Yang(2020)}]{pmlr-v119-sun20c}
Zhiqing Sun and Yiming Yang. 2020.
\newblock \href {http://proceedings.mlr.press/v119/sun20c.html} {An {EM}
  approach to non-autoregressive conditional sequence generation}.
\newblock In \emph{Proceedings of the 37th International Conference on Machine
  Learning}, volume 119 of \emph{Proceedings of Machine Learning Research},
  pages 9249--9258. PMLR.

\bibitem[{Vaswani et~al.(2017)Vaswani, Shazeer, Parmar, Uszkoreit, Jones,
  Gomez, Kaiser, and Polosukhin}]{vaswani2017attention}
Ashish Vaswani, Noam Shazeer, Niki Parmar, Jakob Uszkoreit, Llion Jones,
  Aidan~N. Gomez, Lukasz Kaiser, and Illia Polosukhin. 2017.
\newblock Attention is all you need.
\newblock In \emph{Proceedings of the 31st International Conference on Neural
  Information Processing Systems}, NIPS'17, pages 6000--6010, Red Hook, NY,
  USA. Curran Associates Inc.

\bibitem[{Viterbi(1967)}]{1054010}
A.~Viterbi. 1967.
\newblock \href {https://doi.org/10.1109/TIT.1967.1054010} {Error bounds for
  convolutional codes and an asymptotically optimum decoding algorithm}.
\newblock \emph{IEEE Transactions on Information Theory}, 13(2):260--269.

\bibitem[{Wang et~al.(2018)Wang, Zhang, and Chen}]{wang2018semi}
Chunqi Wang, Ji~Zhang, and Haiqing Chen. 2018.
\newblock \href {https://doi.org/10.18653/v1/D18-1044} {Semi-autoregressive
  neural machine translation}.
\newblock In \emph{Proceedings of the 2018 Conference on Empirical Methods in
  Natural Language Processing}, pages 479--488, Brussels, Belgium. Association
  for Computational Linguistics.

\bibitem[{Wu et~al.(2016)Wu, Schuster, Chen, Le, Norouzi, Macherey, Krikun,
  Cao, Gao, Macherey et~al.}]{wu2016google}
Yonghui Wu, Mike Schuster, Zhifeng Chen, Quoc~V Le, Mohammad Norouzi, Wolfgang
  Macherey, Maxim Krikun, Yuan Cao, Qin Gao, Klaus Macherey, et~al. 2016.
\newblock Google's neural machine translation system: Bridging the gap between
  human and machine translation.
\newblock \emph{arXiv preprint arXiv:1609.08144}.

\bibitem[{Zheng et~al.(2021)Zheng, Zhou, Huang, Chen, Xu, and
  Li}]{zheng2021duplex}
Zaixiang Zheng, Hao Zhou, Shujian Huang, Jiajun Chen, Jingjing Xu, and Lei Li.
  2021.
\newblock \href {http://arxiv.org/abs/2105.03458} {Duplex sequence-to-sequence
  learning for reversible machine translation}.

\bibitem[{Zhou et~al.(2020)Zhou, Gu, and Neubig}]{Zhou2020Understanding}
Chunting Zhou, Jiatao Gu, and Graham Neubig. 2020.
\newblock \href {https://openreview.net/forum?id=BygFVAEKDH} {Understanding
  knowledge distillation in non-autoregressive machine translation}.
\newblock In \emph{International Conference on Learning Representations}.

\end{thebibliography}
\bibliographystyle{acl_natbib}
\end{document}